\documentclass[letterpaper, 10 pt, conference]{ieeeconf}  

\IEEEoverridecommandlockouts                              
\usepackage[utf8]{inputenc}
\usepackage[T1]{fontenc}
\usepackage[backend=biber,style=ieee,sortlocale=de_DE,natbib=true,url=true,doi=false,eprint=false]{biblatex} 
\usepackage[]{todonotes}
\usepackage[destlabel=true,colorlinks=true,urlcolor=blue,linkcolor=blue,citecolor=blue]{hyperref}
\usepackage{soul}
\usepackage[]{float} 
\usepackage{mathtools}
\usepackage{booktabs}

\usepackage{tikz}
\usepackage{relsize}

\usepackage{amsfonts}

\newtheorem{definition}{Definition}

\usepackage{listings}
\usepackage{hyperref}

\usepackage{lipsum}
\usepackage{algorithm}
\usepackage[noend]{algpseudocode}

\usepackage{pgfplots}
\usepgfplotslibrary{groupplots}
\pgfplotsset{compat=1.18}

\usetikzlibrary{patterns,shapes.arrows}
\usetikzlibrary{positioning}
\usetikzlibrary{arrows}
\usetikzlibrary{shapes}
\usetikzlibrary{arrows.meta}
\usepackage{footmisc}

\newcommand{\va}{\mathbf a}

\newcommand{\vf}{\mathbf f}

\newcommand{\vp}{\mathbf p}

\newcommand{\vu}{\mathbf u}
\newcommand{\vv}{\mathbf v}

\newcommand{\vx}{\mathbf x}

\usepackage{subcaption}
\usepackage{placeins}

\title{Hybrid Decision Making for Scalable Multi-Agent Navigation: Integrating Semantic Maps, Discrete Coordination, and Model Predictive Control}

\author{K. de Vos$^{1,*}$, E. Torta$^{1}$, H. Bruyninckx$^{1,2,3}$, C.A. López Martínez$^{1}$, M.J.G. van de Molengraft$^{1}$ 
\thanks{$^{1}$ Department of Mechanical Engineering, Eindhoven University of Technology, The Netherlands}%
\thanks{$^{2}$Department of Mechanical Engineering, KU Leuven, Belgium}
\thanks{$^{3}$Flanders Make, Leuven, Belgium}
\thanks{$^{*}$E-Mail: {\tt\small k.d.vos at tue.nl}}
}

\begin{document}

\maketitle
\thispagestyle{empty}
\pagestyle{empty}

\begin{abstract}

This paper presents a framework for multi-agent navigation in structured but dynamic environments, integrating three key components: a shared semantic map encoding metric and semantic environmental knowledge, a claim policy for coordinating access to areas within the environment, and a Model Predictive Controller for generating motion trajectories that respect environmental and coordination constraints.
The main advantages of this approach include: (i) enforcing area occupancy constraints derived from specific task requirements; (ii) enhancing computational scalability by eliminating the need for collision avoidance constraints between robotic agents; and (iii) the ability to anticipate and avoid deadlocks between agents. The paper includes both simulations and physical experiments demonstrating the framework's effectiveness in various representative scenarios. 
\end{abstract}

\section{Introduction}
\begin{figure}[t]
    \centering
    \begin{subfigure}[t]{0.475\linewidth}
        \centering
        \resizebox{!}{0.075\paperheight}{\definecolor{ao(english)}{rgb}{0.0, 0.5, 0.0}
\definecolor{battleshipgrey}{rgb}{0.21, 0.27, 0.31}
\begin{tikzpicture}[align=center,node distance=1.5cm]
    \draw  [dashed, draw=none] (-3.5,-0.85) node (v6) {} rectangle (-2.5,1.95) node (v2) {} node[pos=0.5] {};
    \draw  [dashed, draw=none] (-3.5,1.95) node (v1) {} rectangle (-2.5,3) node (v3) {} node[pos=0.5] {};
    \draw  [dashed, draw=none] (-2.5,3) rectangle (-0.25,2) node (v5) {} node[pos=0.5] {};
    \draw  [dashed, draw=none] (-5.5,3) rectangle (-3.5,1.95) node [pos=0.5] (v10) {};

    \draw  [fill = {rgb:black,1; white,10}, draw = black] (-3.4,-0.85) rectangle (-3.6,1.75);
    \draw  [fill = {rgb:black,1; white,10}, draw = black] (-2.6,3.25) rectangle (-0.35,3.05) node (v4) {};
    \draw  [fill = {rgb:black,1; white,10}, draw = black] (-5.5,3.25) node (v8) {} rectangle (-3.4,3.05) node (v18) {};
    \draw  [fill = {rgb:black,1; white,10}, draw = black] (-2.6,-0.85) node (v7) {} rectangle (-2.4,1.95);
    \draw  [fill = {rgb:black,1; white,10}, draw = black] (-0.35,1.75) node (v9) {} rectangle (-2.6,1.95) node (v20) {};
    \draw  [fill = {rgb:black,1; white,10}, draw = black] (-3.4,1.75) rectangle (-5.5,1.95);
    \draw  [fill = {rgb:black,1; white,10}, draw = black] (-2.6,3.05) node (v21) {} rectangle (-3.4,3.25);

\draw  [fill=blue,opacity=0.75, draw=none](-3,-0.15) node (v11) {} ellipse (0.2 and 0.2);

\draw  [fill = {rgb:black,1; white,10}, draw = black] (v8) rectangle (-5.3,1.75);
\draw  [fill = {rgb:black,1; white,10}, draw = black](-3.6,-0.7) rectangle (-2.4,-0.85);
\draw [fill = {rgb:black,1; white,10}, draw = black] (-0.55,3.25) rectangle (v9);

\draw [draw= none, fill=red, opacity=0.75] (-2.15,2.75) node (v12) {} ellipse (0.2 and 0.2);
\draw  [draw= none, fill=red, opacity=0.35]  (-2.45,2.75) ellipse (0.2 and 0.2);
\draw  [draw=none, fill=red, opacity=0.25](-2.65,2.75) ellipse (0.2 and 0.2);
\draw  [draw=none, fill=red, opacity=0.2] (-2.85,2.65) ellipse (0.2 and 0.2);
\draw  [draw=none, fill=red, opacity=0.25] (-2.95,2.5) ellipse (0.2 and 0.2);
\draw   [draw=none, fill=red, opacity=0.2](-3,2.3) ellipse (0.2 and 0.2);
\draw   [draw=none, fill=red, opacity=0.15](-3,2.05) ellipse (0.2 and 0.2);

\draw  [draw= none, fill=blue, opacity=0.35](-3,0.1) ellipse (0.2 and 0.2);
\draw  [draw= none, fill=blue, opacity=0.3](-3,0.3) ellipse (0.2 and 0.2);
\draw  [draw= none, fill=blue, opacity=0.25] (-3,0.5) ellipse (0.2 and 0.2);
\draw  [draw= none, fill=blue, opacity=0.2] (-3,0.7) ellipse (0.2 and 0.2);
\draw  [draw= none, fill=blue, opacity=0.15] (-3,0.9) ellipse (0.2 and 0.2);

\draw  [draw= none, fill=ao(english), opacity=0.75] (-5.05,2.2) node (v13) {} ellipse (0.2 and 0.2);
\draw  [draw= none, fill=ao(english), opacity=0.35](-4.8,2.2) ellipse (0.2 and 0.2);
\draw  [draw= none, fill=ao(english), opacity=0.3](-4.55,2.2) ellipse (0.2 and 0.2);
\draw  [draw= none, fill=ao(english), opacity=0.25](-4.3,2.2) ellipse (0.2 and 0.2);
\draw  [draw= none, fill=ao(english), opacity=0.2](-4,2.2) ellipse (0.2 and 0.2);
\draw  [draw= none, fill=ao(english), opacity=0.15](-3.8,2.2) ellipse (0.2 and 0.2);
\node (v14) at (-5.3,2.5) {};
\node (v15) at (-3.4,2.5) {};

\draw [dashed, thick, opacity=0.3] (v14.center) edge (v15.center);
\node (v16) at (-2.6,2.5) {};
\node (v17) at (-0.55,2.5) {};
\draw  [dashed, thick, opacity=0.3] (v16.center) edge (v17.center);
\node (v19) at (-3.4,1.95) {};
    \draw [dashed, thick, opacity=0.3] (v18.center) edge (v19.center);
\draw [dashed, thick, opacity=0.3] (v19.center) edge (v20.center);
\draw [dashed, thick, opacity=0.3] (v20.center) edge (v21.center);

\draw [fill=battleshipgrey, draw=battleshipgrey]  (-3.125,-0.22) rectangle (-3.15,-0.1);
\draw [fill=battleshipgrey, draw=battleshipgrey]   (-2.875,-0.22) rectangle (-2.85,-0.1);
\draw [fill=battleshipgrey, draw=battleshipgrey]  (-3,-0.025) ellipse (0.02 and 0.02);

\draw  [fill=battleshipgrey, draw=battleshipgrey]  (-4.925,2.2) ellipse (0.02 and 0.02);
\draw [fill=battleshipgrey, draw=battleshipgrey]   (-4.98, 2.075) rectangle (-5.1, 2.05);
\draw [fill=battleshipgrey, draw=battleshipgrey]   (-4.98, 2.325) rectangle (-5.1, 2.35);

\draw  [fill=battleshipgrey, draw=battleshipgrey]  (-2.275,2.75) ellipse (0.02 and 0.02);
\draw [fill=battleshipgrey, draw=battleshipgrey]   (-2.22, 2.625) rectangle (-2.1, 2.6);
\draw [fill=battleshipgrey, draw=battleshipgrey]   (-2.22, 2.875) rectangle (-2.1, 2.9);

\end{tikzpicture}}
        \caption{A deadlock would occur if the red agent enters the corridor before the blue agent has left it, as the corridor is too narrow for them to pass each other.}
        \label{fig:deadlock}
    \end{subfigure}
    ~
    \begin{subfigure}[t]{0.475\linewidth}
        \centering        
        \resizebox{\linewidth}{!}{\rotatebox{0}{\input{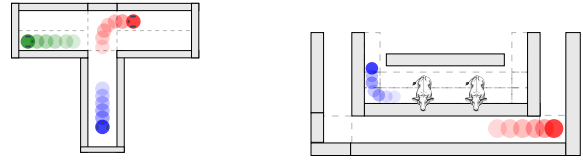}}}
        \caption{Unsafe, or uncomfortable, situations could occur if the areas on both sides of the divider are accessed at the same time.}
        \label{fig:cleanscen}
    \end{subfigure}
    \caption{ Motivating examples for the presented approach }
    \label{fig:motivating}
\end{figure}

The autonomous operation of multiple mobile robots in man-made environments, such as warehousing, manufacturing, and farming, demands motion generation and coordination methods that are safe, robust, efficient and reactive to changes in the environment. Even though these environments are structured with features like aisles, corridors, or rooms, they become dynamic due to the movement of the robotic agents themselves. The environments may even be shared with humans.
Model Predictive Control is a motion control approach that solves a constrained optimisation problem at every sample time, and applies the control input in a receding horizon fashion. 
While MPC has been shown to be effective in the considered settings, e.g. \cite{OMGtools2017, ferranti2018coordination}, scalability problems in terms of computation time and existence of solutions are known to affect the performance of MPC when applied to multi-robot coordination \cite{serra2020whom, SALIMILAFMEJANI2021103774}. In previous work \cite{devos2023automatic}, we presented a method to dynamically configure Model Predictive Controllers (MPC) to solve local navigation problems, in structured environments, based on the information represented in a semantic graph world model.  We, furthermore, showed that, in certain scenarios, the online decision making on which agents should collaborate resulted in a significant reduction of the computation time of the MPCs.
However, within our previous work, the trajectories of coordinating groups of agents were jointly and centrally computed. Which limited the scalability of the approach in, for instance, busy intersection scenarios. Furthermore, in tight environments due to the limited control horizon of the MPCs, deadlocks could occur, for example see Figure \ref{fig:deadlock}. The red and blue agents could get into a deadlock, since the narrow corridor doesn't allow the red agent to pass the blue agent.
Furthermore, our previously proposed method lacked the ability to limit the maximum occupancy of specific areas. This feature may be necessary in scenarios like the one shown in Figure \ref{fig:cleanscen}, where at most one agent must be present in zones on both sides of a divider. For instance, when deploying cleaning and feeding robots in farms, the action of cleaning a zone should not be performed while feeding is taking place on the other side of a feeding fence.

This works extends the work presented in \cite{devos2023automatic} and addresses the aforementioned shortcomings. In particular, we decouple the continuous motion planning of the agents from the coordination between them through the introduction of a discrete claim policy which dynamically configures the MPC controller of each agent. The advantages of this approach are three-fold. Firstly we are able to enforce occupancy constraints which could result from the the task specification of the robot, for example the feeding cleaning scenario in Figure \ref{fig:cleanscen}. Secondly the agents no longer need to incorporate no-collision constraints and optimise their trajectories w.r.t each other since the areas which they reserve are physically disjoint. Thirdly, by introducing a claim policy that manages the access to areas we are able to detect and avoid deadlock situations. Our Contributions are threefold. Firstly we propose a distributed MPC approach with coordination resolved at discrete level, which is dynamically reconfigured at run time. Secondly we extend the graph-based WM from~\cite{devos2023automatic} from which the MPC, and discrete coordination algorithms, can be automatically configured. Thirdly we present a framework to enforce task-dependent maximum occupancy constraints on semantic regions.\\
\textit{Related Work}
A wide range of approaches for navigation and motion planning have been studied within the context of multi-agent robotic systems. Planning-based methods assume sufficient knowledge of the system is available to plan the robot actions offline (see \cite{standley2011complete,WAGNER20151, 7138650} for examples). When significant deviations occur in agent or environment behavior, plans need to be (partially) recomputed. In contrast, reactive approaches (see \cite{580977, fiorini1998motion, 4543489} for examples) use locally available sensor data to rapidly change actions based on acquired information, often at the expense of global optimality and deadlock or livelock freeness.
Within Model Predicitve control (MPC) approches control inputs are obtained from solving a constrained optimisation problem at every sample time. The objective function is a measure of the desired performance of the system. The constraints arise from the system dynamics, hardware limits, and (safety-) requirements. MPC controllers predict future states over a pre-specified horizon, combining the advantages of online reactivity with the longer prediction horizons of planning methods.
\noindent
This paper considers motion planning for autonomous mobile robots in structured environments, the authors of \cite{ferranti2018coordination} similarly considered the navigation of multiple autonomous vessels at canal intersections in their design of a distributed nonlinear MPC, in which decisions on vehicle priority directly result from the solution of the decentralized MPC problem.
The authors of \cite{OMGtools2016} develop an MPC and employ the separating hyperplane constraint formulation to describe the no-collison constraints for a mobile robot navigating in a environment containing convex polygonal obstacles.
%
%
However, as stated in, among others, \cite{SALIMILAFMEJANI2021103774, serra2020whom} the computational cost of MPC approaches grows large for a large number of agents. Employing distributed MPC control approaches, like in \citep{ferranti2018coordination, OMGtools2017, serra2020whom}, reduces this computational cost, however it is challenging to guarantee deadlock-free behaviour in these distributed settings.
\noindent
Similar to the work in \cite{10292991}, the framework proposed in this paper combines continuous methods for the motion control of the agent, in particular MPC, with discrete methods, for high-level collision and deadlock avoidance. As stated in \cite{10292991}, the discrete control component is effective at collision and deadlock avoidance and, furthermore, allows for a reduction in the scale of the optimization problems which constitute the continuous control component. The authors of \cite{roszkowska2023multi} similarly propose a three-tier decision-making framework consisting of cell-transition control using discrete event systems, cell-traverse control with a continuous-time formulation, and robot velocity control. Their approach considers independently planned robot paths. Coordination is achieved through real-time refinement to prevent collisions and deadlocks. A supervisor centrally controls robot motion between discrete stages of the robot plans. Accessibility of spaces, and deadlock avoidance, are considered within the context of the Resource Allocation System framework. In contrast to the approach presented in this paper, local navigation is accomplished via path following. Collision avoidance, necessary when multiple agents occupy the same cell, is managed through roundabout control and Artificial Potential Fields. The authors of \cite{alonso2018reactive} combine a distributed local motion planner  with guarantees on collision avoidance, specifically Reciprocal Velocity Obstacles, with a centralized automaton which is synthesised offline from a mission specification in Linear Temporal Logic to achieve guaranteed collision avoidance and the ability to resolve deadlocks. Our method, instead of relying on offline synthesis, is online configurable and adaptable through the semantic world model.

\section{Method}

\subsection{MPC Formulation}
\label{sec:control_layer}
The local navigation problem is formulated as an MPC problem. However, opposed to \cite{devos2023automatic} the MPC is formulated, and solved, per agent i.s.o per group of agents.
Consider a set of objectives $\mathbb{O}_i$ and a set of area boundaries, walls and obstacles $\mathbb{W}_i$, for each agent $A_i$ the single-agent local navigation MPC problem can be formulated as:
\begin{align}
     & \min_{\vx (\cdot), \vu(\cdot)} \sum_{k_t=0}^{N_t} l(\vx, \vu, k_t) + n(\vx) , \label{eq:obj}
\end{align}
\noindent
Subject to:
\begin{align}
    \quad & \vx(t + k_t + 1) = \vf(\vx( t + k_t ), \vu( t + k_t)),                                                               \\
          & \vx(t) = \vx_{init},                                                                                        \\
          & \vx_{min} \leq \vx( t  + k_t ) \leq \vx_{max},                                                                   \\
          & \vu_{min} \leq \vu( t + k_t) \leq \vu_{max},                                                                    \\
          & \text{dist}\left(A\left( t + k_t \right), w\right) > 0, \quad  \forall w \in \mathbb{W}, \label{eq:no_col_wall}
\end{align}
in which $\vf_i$ describes the dynamic model of agent $a_i$. Furthermore, $\vx_i \in \mathbb{X}_i$ and $\vu_i \in \mathbb{U}_i$ represent the state and input of agent $a_i$ respectively. With $l(\vx,\vu,k,k_t)$ and $n(\vx)$ the stage and terminal cost functions respectively. The state of agent $a_i$ describes its longitudinal velocity and its Cartesian position and orientation w.r.t a global reference frame. The control inputs of agent $a_i$ are the acceleration in longitudinal direction and its angular velocity. 
Utilizing the separating hyperplane constraint formulation as described in \cite{OMGtools2016, 7810517}, the no-collision constraints \eqref{eq:no_col_wall} , for each element  $w \in \mathbb{W}_i$ with vertices $\vv_{w}$ for a circular~agent $A_i$ with radius~$r_v$~as, are formulated as:
\begin{align}
    \va_{w}^T \vv_{w,j}- b_{w} & \geq 0,    &  & \forall \vv_{w} \in \vv_w, \\
    \va_{w}^T \vx(t + k_t) - b_{o}   & \leq -r_v, &  & \forall k_t \in [0 , N_t],     \\
    ||\va_{w}||_2              & \leq 1,
\end{align}
with $\va$ and $b$ the normal vector to and offset of the hyperplane respectively.

\subsection{Semantic World Model}
\label{sec:world_model}
The semantic map is represented as a graph that models the environment in which the agents are deployed on a topological level. An identical semantic map of the environment is made available to all agents, and assumed to be correct and up-to-date with the current state of said environment. The semantic map is thus defined as:

\begin{definition}[Semantic Map]\label{def:semmap}
    a graph $G=(V,E)$, with vertices $V$ of type \{Area, Boundary, Interface\}, and edges $E$ between vertices that model the relationships between them, to model the additional properties of the entities.
\end{definition}
The area, boundary and Interface semantic map primitives identified in Definition \ref{def:semmap} can be further defined as:\\
     \textbf{Area}: Areas model the drivable space that is available to agents. Areas can represent physical spaces, such as rooms and corridors, or can be defined virtually. Areas can, furthermore, be contained within super-areas, or contain sub-areas, as is shown in Figure \ref{fig:semi_struc_enva} and \ref{fig:semi_struc_envb}. Figure \ref{fig:semi_struc_enva} shows a simple structured environment, in which the (super-)area $S_0$ is composed of the individual sub-areas $S_{00}$ and $S_{01}$, which represent separate driving lanes. Figure \ref{fig:semi_struc_envb} separately shows the super- and sub-area views of this area. Figure \ref{fig:semi_struc_envc} shows an excerpt of the corresponding graph model, i.e. the semantic map.
     \textbf{Boundary}:
          In the semantic map the limits of areas are modeled through boundaries. A physical interpretation of this primitive are the walls that line a certain area, such as a corridor. However, within the semantic map, boundaries do not necessarily correspond to physical elements, and can also be defined virtually. Boundaries are treated as constraints in the MPC formulation, and are members of the set $\mathbb{W}$ in \eqref{eq:no_col_wall}.  
     \textbf{Interface}:
          Interfaces model the connectivity between areas. For each agent, given their route, interfaces are classified at runtime into two types: \textit{active} and \textit{inactive}.
          Active interfaces are only those interfaces which connect the areas \textit{the agent is allowed to access} along the route of the agent. The remaining interfaces are \textit{inactive}. \textit{Active} interfaces are used to configure the objective \eqref{eq:obj} of the agents. \textit{Inactive} interfaces are treated as virtual boundaries, i.e. are added to $\mathbb{W}$ in \eqref{eq:no_col_wall}. This prevents the agent from entering areas that are either outside of its route or which the agent is not allowed to access. For example, an agent deployed in the environment sketched in Figure \ref{fig:semi_struc_envd} with route $[ S_3, S_1, S_2]$ would consider interfaces $I_4$ and $I_2$ active, and interfaces $I_5$, $I_3$, and $I_1$ as inactive.
\begin{figure}[t]
    \centering
    \resizebox{!}{0.08\paperheight}{\begin{tikzpicture}[align=center,node distance=1.5cm]
    \draw  [dashed] (-3.5,0.55) node (v6) {} rectangle (-2.5,2) node (v2) {} node[pos=0.5] {$S_0$};
    \draw  [dashed] (-3.5,2) node (v1) {} rectangle (-2.5,3) node (v3) {} node[pos=0.5] {$S_1$};
    \draw  [dashed] (-2.5,3) rectangle (-0.35,2) node (v5) {} node[pos=0.5] {$S_2$};
    \draw  [dashed] (-5.5,3) rectangle (-3.5,2) node [pos=0.5] (v10) {$S_3$};

    \draw  [fill = {rgb:black,1; white,10}, draw = black] (-3.55,0.55) rectangle (-3.75,1.75);
    \draw  [fill = {rgb:black,1; white,10}, draw = black] (-2.5,3.25) rectangle (-0.35,3.05) node (v4) {};
    \draw  [fill = {rgb:black,1; white,10}, draw = black] (-5.5,3.25) rectangle (-3.5,3.05);
    \draw  [fill = {rgb:black,1; white,10}, draw = black] (-2.45,0.55) node (v7) {} rectangle (-2.25,1.95);
    \draw  [fill = {rgb:black,1; white,10}, draw = black] (-0.35,1.75) rectangle (-2.45, 1.95);
    \draw  [fill = {rgb:black,1; white,10}, draw = black] (-3.55,1.75) rectangle (-5.5,1.95);
    \draw  [fill = {rgb:black,1; white,10}, draw = black] (-2.5,3.05) rectangle (-3.5,3.25);
    
    \node (v8) at (-0.35,3) {};
    \draw  [red] (v6.center) -- (v7.center) node [midway, above] {$I_0$};
    \draw  [red] (v1.center) -- (v2.center) node [midway, below] {$I_1$};
    \draw  [red] (v2.center) -- (v3.center) node [midway, right] {$I_2$};
    \draw  [red] (v5.center) -- (v8.center) node [midway, left] {$I_3$};

\node (v9) at (-3.5,3) {};
\node (v11) at (-3.5,2) {};
\draw  [red] (v9.center) -- (v11.center) node [midway, left] {$I_4$};
\node (v12) at (-5.5,3) {};
\node (v13) at (-5.5,2) {};
\draw  [red] (v12.center) -- (v13.center) node [midway, right] {$I_5$};
\end{tikzpicture}}
    \caption{A simplified depiction of a T-junction environment. Perimeters of areas ($S_i$) are drawn in black dashed lines. Boundaries of areas are drawn in gray. Interfaces ($I_j$) between areas are drawn in red.}
    \label{fig:semi_struc_envd}
\end{figure}
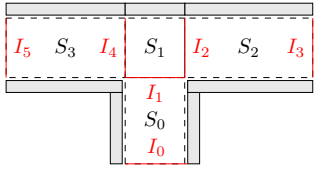
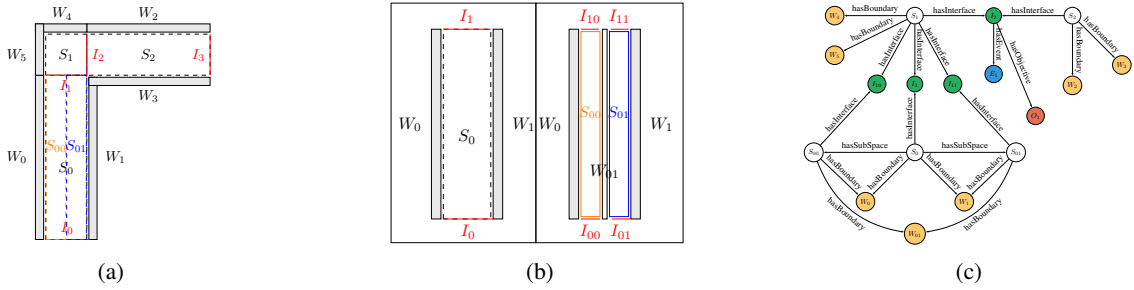
\begin{figure*}[t]
    \centering
    \begin{subfigure}[c]{0.3\linewidth}
        \centering
        \resizebox{!}{0.115\paperheight}{\begin{tikzpicture}[align=center,node distance=1.5cm]
    \draw  [dashed] (-3.5,-2) node (v6) {} rectangle (-2.5,2) node (v2) {} node[pos=0.5, below] {$S_0$};
    \draw  [dashed] (-3.5,2) node (v1) {} rectangle (-2.5,3) node (v3) {} node[pos=0.5] {$S_1$};
    \draw  [dashed] (-2.5,3) rectangle (0.5,2) node (v5) {} node[pos=0.5] {$S_2$};

    \draw  [fill = {rgb:black,1; white,10}, draw = black] (-3.55,-2) rectangle (-3.75,2);
    \draw  [fill = {rgb:black,1; white,10}, draw = black] (-2.5,3.25) rectangle (0.5,3.05) node (v4) {};
    \draw  [fill = {rgb:black,1; white,10}, draw = black] (-2.45 , -2) node (v7) {} rectangle (-2.25,1.95);
    \draw  [fill = {rgb:black,1; white,10}, draw = black] (0.5, 1.75) rectangle (-2.45, 1.95);
    \draw  [fill = {rgb:black,1; white,10}, draw = black] (-2.5,3.05) rectangle (-3.75,3.25);
     \draw  [fill = {rgb:black,1; white,10}, draw = black] (-3.55,2) rectangle (-3.75,3.25);
    
    \node (v8) at (0.5,3) {};
    \draw  [red] (v6.center) -- (v7.center) node [midway, above] {$I_0$};
    \draw  [red] (v1.center) -- (v2.center) node [midway, below] {$I_1$};
    \draw  [red] (v2.center) -- (v3.center) node [midway, right] {$I_2$};
    \draw  [red] (v5.center) -- (v8.center) node [midway, left] {$I_3$};

\node at (-1,1.5) {$W_3$};
\node (v9) at (-4.2,0) {$W_0$};
\node at (-1,3.5) {$W_2$};
\node at (-1.8,0) {$W_1$};
\node at (-3.1,3.5) {$W_4$};
\node at (-4.2,2.5) {$W_5$};

\draw  [dashed, orange]  (-3.49,-2) rectangle (-3.01,2) node (v10) {} node[pos=0.5, above] {$S_{00}$};
\draw  [dashed, blue]  (-2.99 ,2)  rectangle  (-2.51,-2) node (v11) {} node[pos=0.5, above] {$S_{01}$};

\draw  (-1.25,-1.35) rectangle (-1.25,-1.35);
\end{tikzpicture}}
        \caption{}
        \label{fig:semi_struc_enva}
    \end{subfigure}
    ~
    \begin{subfigure}[c]{0.3\linewidth}
        \centering
        \resizebox{!}{0.115\paperheight}{\begin{tikzpicture}[align=center,node distance=1.5cm]
    \draw  [dashed] (-3.5,-2) node (v60) {} rectangle (-2.5,2) node (v20) {} node[pos=0.5, below] {$S_0$};
    
    \node(v10) at  (-3.5,2) {};
    
    \draw  [fill = {rgb:black,1; white,10}, draw = black] (-3.55,-2) rectangle (-3.75,2);
    \draw  [fill = {rgb:black,1; white,10}, draw = black] (-2.45 , -2) node (v70) {} rectangle (-2.25,2);

   \node(v7) at (-2.45 , -2){};
    \draw  [red] (v60.center) -- (v70.center) node [midway, below] {$I_0$};
    \draw  [red] (v10.center) -- (v20.center) node [midway, above] {$I_1$};

\node (v90) at (-4.2,0) {$W_0$};
\node at (-1.8,0) {$W_1$};

    \draw  [fill = {rgb:black,1; white,10}, draw = black] (-0.65,-2) rectangle (-0.85,2);
    \draw  [fill = {rgb:black,1; white,10}, draw = black] (0.45,-2) node (v7) {} rectangle (0.65,2);
    \draw  [fill = {rgb:black,1; white,45}, draw = black] (-0.05,-2) node (v7) {} rectangle (-0.15,2);
      
    \node(v1) at (-0.6,2) {};
    \node(v2) at (0.4,2) {};
    \node(v6) at (-0.6,-2) {};

   \node(v7) at (0.45,-2) {};
    \draw  [draw=none] (v6.center) -- (v7.center) node (v13) [midway, below] {};
    \draw  [red] (v6.center) -- (v13.north west) node [midway, below] {$I_{00}$};
    \draw  [red] (v7.center) -- (v13.north east) node [midway, below] {$I_{01}$};
    \draw  [draw=none] (v1.center) -- (v2.center) node (v14) [midway, above] {};
    \draw  [red] (v1.center) -- (v14.south west) node [midway, above] {$I_{10}$};
    \draw  [red] (v2.center) -- (v14.south east) node [midway, above] {$I_{11}$};

\draw  [orange]  (-0.6,-1.95) rectangle (-0.2,1.95) node (v10) {} node[pos=0.5, above] {\small $S_{00}$};
\draw  [blue]  (0,1.95)  rectangle  (0.4,-1.95) node (v11) {} node[pos=0.5, above] {\small $S_{01}$};

\node (v9) at (-1.25,0) {$W_0$};
\node at (1.1,0) {$W_1$};
\node (v9) at (-0.1,-1) {$W_{01}$};

\draw  (-4.6,2.55) rectangle (-1.55,-2.5);
\draw  (1.55,2.55) rectangle (-1.55,-2.5);
\end{tikzpicture}}
        \caption{}
        \label{fig:semi_struc_envb}
    \end{subfigure}
    ~
    \begin{subfigure}[c]{0.3\linewidth}
        \centering
        \resizebox{!}{0.115\paperheight}{\begin{tikzpicture}[align=center,node distance=2cm]

\definecolor{myred}{rgb}{0.906,0.298,0.235}
\definecolor{myblue}{rgb}{0.204, 0.596, 0.859}
\definecolor{myyellow}{rgb}{0.992,0.796,0.431}
\definecolor{mygray}{rgb}{0.698, 0.745, 0.765}
\definecolor{mygreen}{rgb}{0.152, 0.682, 0.376}
\definecolor{mypurple}{rgb}{0.698, 0.745, 0.765}
\definecolor{mybrown}{rgb}{0.882, 0.439, 0.333}

\node [circle, draw] (S1) at (0, 0) { $S_1$};
\node [circle, draw, below = 2.5cm of S1, fill = mygreen] (I1) {  $I_1$};
\node [circle, draw, left = 1 cm of I1, fill = mygreen] (I10) {  $I_{10}$};
\node [circle, draw, right =  1 cm of I1, fill = mygreen] (I11) {  $I_{11}$};
\node [circle, draw, right  = 3cm of S1, fill = mygreen] (I2) {  $I_2$};


\node [circle, draw,  below = of I2, fill = myblue] (E1) {  $E_1$};
\node [circle, draw,  below right = of E1,  fill = mybrown] (O1)  {  $O_1$};

\node [circle, draw, below = 2.5cm of I1] (S0) {  $S_0$};
\node [circle, draw, below left  =2.5cm of S0, fill = myyellow] (W0) {  $W_0$};
\node [circle, draw, below right  =2.5cm of S0, fill = myyellow] (W1){  $W_1$};

\node [circle, draw, right   = 3cm of I2] (S2) {  $S_2$};
\node [circle, draw, below = 2.5 cm of S2, fill = myyellow] (W2) {  $W_2$};
\node [circle, draw, below right = 2.5cm of S2, fill = myyellow] (W3) {  $W_3$};

\node [circle, draw, left = 3 cm of S1, fill = myyellow] (W4) {  $W_4$};
\node [circle, draw, below =1 cm of W4, fill = myyellow] (W5) {  $W_5$};

\node [circle, draw, left = 4 cm of S0] (S00) {  $S_{00}$};
\node [circle, draw, right  = 4cm of S0] (S01) {  $S_{01}$};

\node [circle, draw, below =3cm of S0, fill = myyellow] (W01a){  $W_{01}$};

\draw [-latex]  (S1) edge node [sloped, anchor = center, above] {  \large hasInterface} (I2);
\draw [-latex]  (S2) edge node [sloped, anchor = center, above] {  \large hasInterface} (I2);
\draw [-latex]  (S0) edge node [sloped, anchor = center, above] {  \large hasInterface} (I1);
\draw [-latex]  (S1) edge node [sloped, anchor = center, above] {  \large hasInterface} (I1);

\draw [-latex]  (S00) edge node [sloped, anchor = center, above] {  \large hasInterface} (I10);
\draw [-latex]  (S1) edge node [sloped, anchor = center, above] {  \large hasInterface} (I10);

\draw [-latex]  (S01) edge node [sloped, anchor = center, above] {  \large hasInterface} (I11);
\draw [-latex]  (S1) edge node [sloped, anchor = center, above] {  \large hasInterface} (I11);

\draw [-latex]  (S00) edge node [sloped, anchor = center, above] { \large hasBoundary} (W0);
\draw [-latex]  (S01) edge node [sloped, anchor = center, above] {  \large hasBoundary} (W1);
\draw [-latex, bend right]  (S00) edge node [sloped, anchor = center, below] {  \large hasBoundary} (W01a);
\draw [-latex, bend left]  (S01) edge node [sloped, anchor = center, below] {  \large hasBoundary} (W01a);

\draw [-latex]  (S2) edge node [sloped, anchor = center, above] {  \large hasBoundary} (W3);
\draw [-latex]  (S2) edge node [sloped, anchor = center, above] {  \large hasBoundary} (W2);

\draw [-latex]  (S0) edge node [sloped, anchor = center, above] {  \large hasBoundary} (W0);
\draw [-latex]  (S0) edge node [sloped, anchor = center, above] {  \large hasBoundary} (W1);

\draw [-latex]  (S1) edge node [sloped, anchor = center, above] {  \large hasBoundary} (W4);
\draw [-latex]  (S1) edge node [sloped, anchor = center, above] {  \large hasBoundary} (W5);

\draw [-latex]  (S0) edge node [sloped, anchor = center, above] {  \large hasSubSpace} (S01);
\draw [-latex]  (S0) edge node [sloped, anchor = center, above] {  \large hasSubSpace} (S00);


\draw [-latex]  (I2) edge node [sloped, anchor = center, above] {  \large hasObjective} (O1);
\draw [-latex]  (I2) edge node [sloped, anchor = center, above] {  \large hasEvent} (E1);

\draw  (-0.3,0.4) rectangle (-0.3,0.4);
\end{tikzpicture}}
        \caption{}
        \label{fig:semi_struc_envc}    
    \end{subfigure}
    \caption{(a) Simplified structured environment consisting of three areas ($S_0$, $S_1$ and $S_2$), and two sub-areas ($S_{00}$ and $S_{01}$ sub-areas of $S_0$). (b) two different views of the same physical area. (c) Excerpt of a graph world model describing the environment in (a). The colors of the nodes signify their different types.}
    \label{fig:semi_struc_graph}
\end{figure*}
\subsection{Coordination Strategy}
\label{sec:coordination_layer}
Consider a set of agents $\mathbb{A}$, with each agent $a_i \in \mathbb{A}$ having a route $p_i$. The route of each agent $p_i = [ S_{i,0}, \dots S_{i, n} ]$ is an ordered list of areas, as introduced in Definition~\ref{def:semmap}. The index of the current area in the route is denoted $k_t$, and $p_i^{(k_t:)}$ denotes the route from the current index forward. We additionally introduce $\vp_i^{(k)}$ to denote the set containing the area $p_i^{(k)}$ and its sub-areas and $\bar{\vp_i}^{(k)}$ containing the area $p_i^{(k)}$ and its super-areas as modeled in the world model.
For every pair of agents $a_i$, $a_j$ with $i$ unequal to $j$ we define, at run time, a conflict set as $C_{i,j} := \{ \vp_i^{(k_t:)} \} \cap \{ \vp_j^{(k_t:)} \} $. The set $C_{i,j}$ thus contains all future areas to be visited by both agents $i$, and $j$, potentially causing conflicts.\\
\noindent
\textbf{Claim Policy}
To achieve coordination between the set of agents, we propose Algorithm \ref{alg:al1}, and introduce a set of indices, that correspond to stages of a route $p_i$ of an agent $a_i$. To coordinate the access to areas between a set of agents, areas $p^{(k)}_i$ can be claimed by the agents. Areas can only be entered by agents that have claimed an area $p^{(k)}_i$.
We introduce the set of Areas claimed by agent $a_i$ as $\mathbb{R}_i$ and the set of claimed areas by any agent as $\mathbb{R} = \cup_{a_i \in \mathbb{A}} \mathbb{R}_i$. The claim policy proposed within this section aims to satisfy the following requirements:
\begin{enumerate}
    \item Areas can simultaneously be claimed by at most one agent: $$ p^{(k)}_i \in \mathbb{R}_j \rightarrow \big( \forall a_j \in \mathbb{A}_j:\; i \neq j: \; \vp^{(k)}_i \cap \mathbb{R}_j = \emptyset \big) .$$

    \item As long as the agents have not finished execution of their route, there should always be at least one agent which can progress:
          $$\exists a_i \in \mathbb{A} \; ( |\mathbb{R}_i| > 1 ) \quad \lor \quad \forall a_i \in \mathbb{A} \; (k_t = k_{\text{max}}).$$
\end{enumerate}
For this, it is assumed that the routes $p_i$ of agents $a_i \in \mathbb{A}$ have distinct starting areas, distinct final areas, and that there exists a trajectory, of areas, that simultaneously satisfy the route $p_i$ for each agent $a_i$ and the requirements on the claim policy. 
For example, we do not consider the case in which two agents are deployed in the environment in Figure \ref{fig:semi_struc_envd}, and in which the agents have routes $[S_3, S_1, S_2]$ and $[S_2, S_1, S_3]$ respectively. Since no trajectory exists which satisfies the route and both requirements of the claim policy. \label{sec:claimPolicy}\\
\noindent
\begin{algorithm}
    \caption{Claim Policy}
    \label{alg:al1}
    \begin{algorithmic}[1]
        \State \textbf{Initialization:}
        \State Initialize: $\mathbb{R}$
        \State Compute: $C_{i,j}$ for $a_i, a_j \in \mathbb{A}$, $i\neq j$
    \State \textbf{Claim current area:}
        \For{all agents $a_i \in \mathbb{A}$}
            \State $\vp_i^{(k_t)} \rightarrow \mathbb{R}_i$
        \EndFor
    \State \textbf{Claim conflict areas}
        \For{all agents $a_i \in \mathbb{A}$}
                \State Compute: $k_{free}$, given: $\vp_i, \; C_{i,j}$   
                \State Compute: $k_{claim}$, given: $\vp_i, \; \mathbb{R}$
                \State Compute: $k_{nblock}$, given: $\vp_i, \; \vp_j, \; \mathbb{R}$
                \State
                \If{$k_{claim} \geq k_{free}$}
                    \State $\vp_i^{([k_t + 1 : k_{free}])} \rightarrow \mathbb{R}_i$
                \Else{ }
                    \State $\vp_i^{([k_t + 1 : k_{nblock}])} \rightarrow \mathbb{R}_i$
                \EndIf
        \EndFor
    \State \textbf{Claim remaining non-conflict areas:}
        \For{all agents $a_i \in \mathbb{A}$}
            \State Compute: $k_{nconf}$, given: $\vp_i, \; C_{i,j}$  
            \State $\vp_i^{([k_t + 1 : k_{nconf}])} \rightarrow \mathbb{R}_i$
        \EndFor
    \end{algorithmic}
\end{algorithm}
    \noindent
    In line 2, of algorithm \ref{alg:al1}, the set of claimed areas is initialised to the empty set.
    Agents claim their current area, i.e. the area with index $k_t$, in Line 6.
    Line 9 computes $k_{free}$, the smallest index that corresponds to an area that an agent can wait in without blocking any other agent, i.e. the smallest index $\kappa$ which satisfies:
    $$ (\forall a_j \in \mathbb{A} \; \land \; \kappa> k_t: \; \; \vp_i^{(\kappa)} \cap C_{i,j} = \emptyset) \lor (\kappa = |p_i| + 1).$$
    Line 10 computes $k_{claim}$, the largest index of the route that can be claimed, i.e. the largest index $\kappa$ which satisfies:
    $$ \forall k: \; k_t \leq k \leq \kappa: \; \vp_i^{(k)} \cap \mathbb{R}_{j} = \emptyset \; \; \forall a_j \in \mathbb{A}.$$ \noindent
   The index $k_{nblock}$ is computed in Line 11, it is the largest index which is claimable and in which the agent $a_i$ doesn't directly cause a deadlock with agent $a_j$ by claiming that area, i.e. the largest index $\kappa$ that satisfies:
    $$
    \kappa \leq k_{free} \ \land \kappa \leq k_{claim} \ \land p_i^\kappa \notin \mathbb{T} \ \land
    $$
     $$ \big(\forall p^q_j, \; p^q_j \in \bar{\vp}^\kappa_i \; \rightarrow \;  \vp_i^{\kappa+1} \cap \vp_j^{q-1} = \emptyset \big). 
    $$
    In which the set $\mathbb{T}$ is the set of areas which the agent can't claim due to them being \textit{in conflict} goal areas. This set is defined as the set corresponding to the largest value of $\kappa$:
    $$
        \mathbb{T} := \{ p^k_i :\forall 0\leq k\leq\kappa, \ \forall a_j \in \mathbb{A};
    $$
    $$
        \vp_i^{(n-k)} \cap \vp_j^{(m-k)} \neq \emptyset\}.
    $$
    in which $\vp_i^{(n)}$ is the final area of agent $a_i$.
    Line 19 computes $k_{nconf}$ which corresponds to the non-conflict areas we can claim ahead within a certain horizon length, i.e. the largest index $\kappa$ that satisfies:
    $$ \forall k: \; k_t \leq k \leq \kappa: ( \vp_i^{(k)} \cap C_{i,j} = \emptyset  \; \forall a_j \in \mathbb{A} $$ 
    $$\lor \; \vp_i^{(k)} \cap \mathbb{R}_i \neq \emptyset )\; \land \; \kappa \leq N_{ch} .$$
\noindent

\noindent
\textbf{Coordinating agent selection}
\label{sec:coordagent}
Whenever a pair of agents $a_i, a_j$ satisfy $C_{i,j} \neq \emptyset$ there might be a need for coordinating the actions of those agents. However, whenever conflict resolution can be postponed, coordination might not (yet) be necessary. From the claim policy outlined in Algorithm \ref{alg:al1} the minimum extent of this horizon is up to index $k_{free}$, i.e. the next area which is not contained in any other agent's route.
We thus define the extent of the horizon as: 
$k_{h} := \text{max} \left( k_{free}, k_t  + N_{h} \right)$, to ensure a constant minimum  length of $N_h$.
Which thus results in the sub-route $\vp^{c}_{i} := \vp_i^{(k_t:k_{h)}}$. When the coordination horizons $\vp^{c}_{i}, \vp^{c}_{j}$ of agents $a_i, a_j \in \mathbb{A}$ overlap, i.e. $\{\vp^{c}_{i}\} \cap \{\vp^{c}_{j}\} \neq \emptyset$, the pair $(a_i, a_j)$ is added to the list of agent pairs requiring coordination. This list serves as the input to Algorithm 2 in \cite{devos2023automatic} which outputs groups of agents needing coordination. These groups then coordinate their actions by adhering to the claim policy introduced in Algorithm \ref{alg:al1}. 
\noindent

\subsection{Controller Configuration}
At run-time, during initialisation of the algorithm, a number of consecutive steps are taken. In the first step the route of each agent is analyzed in order to determine the current need for coordination between agents, according to the procedure described in Section \ref{sec:coordagent}.\\
Within each group of coordinating agents, Algorithm \ref{alg:al1} determines the area claims for each agent within the group.\\
Each agent configures its MPC controller, based on the claims resulting from Algorithm \ref{alg:al1}, retrieving the relevant knowledge and geometry from the world model. More specifically, interfaces to reserved areas are made active, and accounted for in the objective function \eqref{eq:obj}, interfaces to non-reserved areas are made inactive and added to  $\mathbb{W}$ in \eqref{eq:no_col_wall}. Additionally, the MPC controllers query the world model to retrieve the boundaries, objectives, and other relevant information about the areas they are configured to traverse. After initialisation, every time an agent crosses an interface, the groups of coordinating agents, and area reservations are recomputed, and the MPC controllers are reconfigured accordingly.\\
Then at every sample time we predict the active area for each future time step in the agent's control horizon. We update the area-dependent objective and constraints, e.g. distance to the end of the area and velocity limits respectively, accordingly.

\section{Validation}

\subsection{Implementation}
The framework presented in this paper was implemented in Python. The MPC controllers are implemented using the do\_mpc package \cite{DOMPC2017}, which is based on Casadi \cite{CasADi2019}. The problems are solved using the HSL\_MA57 solver \cite{HSL} in IPOPT \cite{wachter2006implementation}. The graph world model is made accessible via a graphDB server. The provided algorithms use the SPARQL query methods in the RDFLib  \cite{RDFLib} package to retrieve the required data from the server. Experiments are performed both in PyBullet \cite{coumans2021} simulation and on the ROSbot 2 Pro mobile robot platform.

\begin{figure}[t]
    \centering
    \begin{subfigure}[c]{0.9\linewidth}
        \centering
        \includegraphics[height=0.095\paperheight]{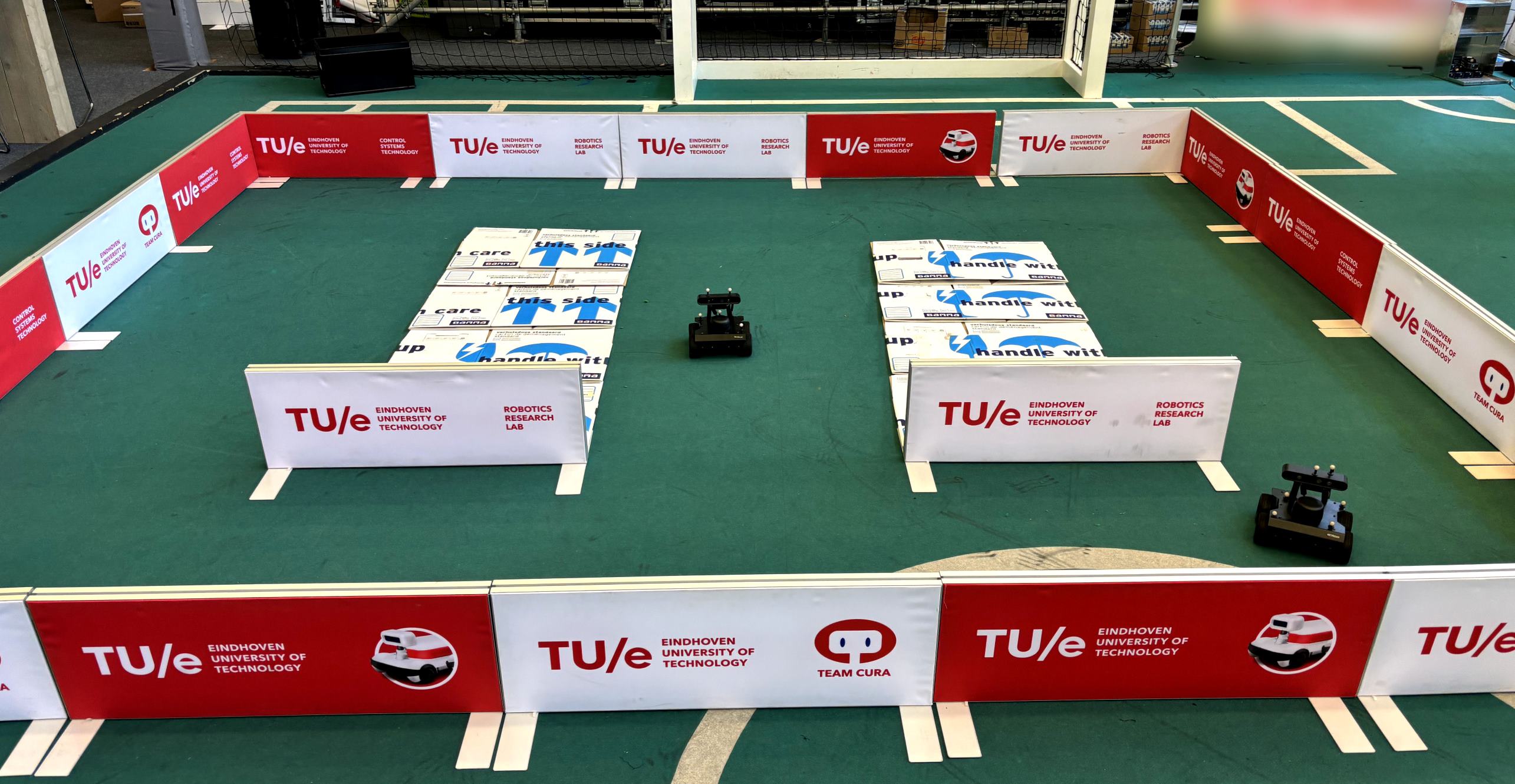}
        \caption{Experimental lab setup with two ROSbot 2 PRO mobile robots}
        \label{fig:experimentsetup}
    \end{subfigure}

    \begin{subfigure}[c]{0.9\linewidth}
        \centering
        \includegraphics[height=0.11\paperheight]{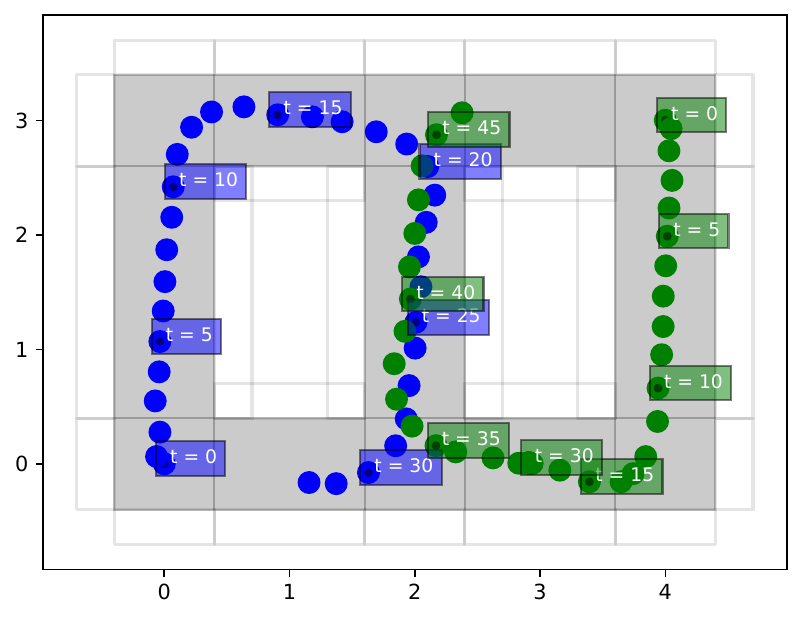}
        \label{fig:trajectoryresults}
        \caption{Trajectory of robotic agents in one of the physical experiments}
    \end{subfigure}
    
    \caption{Experimental validation on physical robotic hardware}
    \label{fig:experimentalresults}
\end{figure}

\subsection{Validation Scenarios}

\begin{figure*}[t]
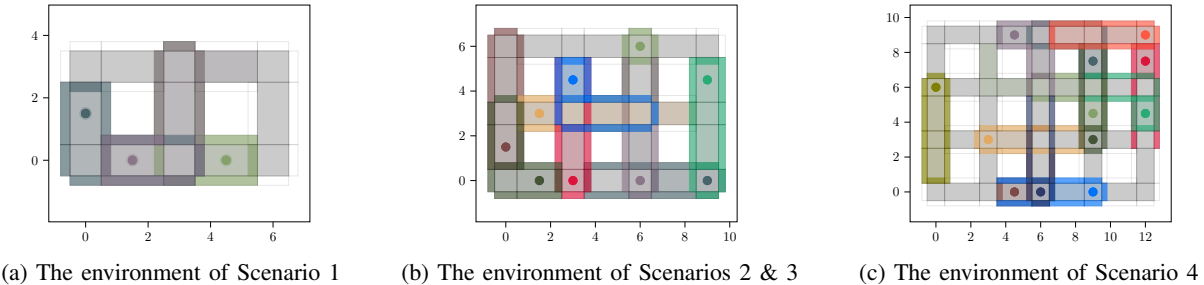

    \centering
    \begin{subfigure}[t]{0.3\linewidth}
        \centering
        \resizebox{!}{0.6\linewidth}{\input{Figures/scen1}}
        \caption{The environment of Scenario 1}
        \label{fig:scen1}
    \end{subfigure}
    ~
    \begin{subfigure}[t]{0.3\linewidth}
        \centering
        \resizebox{!}{0.6\linewidth}{\input{Figures/scen3}}
        \caption{The environment of Scenarios 2 \& 3}
        \label{fig:scen23}
    \end{subfigure}
    ~
    \begin{subfigure}[t]{0.3\linewidth}
        \centering
        \resizebox{!}{0.6\linewidth}{\input{Figures/scen4}}
        \caption{The environment of Scenario 4}
        \label{fig:scen4}
    \end{subfigure}
    \caption{Agents deployed in the validation scenario environments. The grey areas signify drivable space. The colored polygons signify the considered constraints by the respective agents}
    \label{fig:motivating}
\end{figure*}

Four validation scenarios, with increasing complexity, are introduced. In the first scenario, three  agents are deployed in the environment depicted in Figure \ref{fig:scen1}. The second and third scenarios deploy six and nine agents, respectively, in the environment shown in Figure \ref{fig:scen23}. Finally the fourth scenario deploys 12 agents in the environment depicted in Figure \ref{fig:scen4}.
For each validation scenarios we perform 50 simulations\footnote{Because of the extended simulation times, the validation of method~\cite{devos2023automatic} scenario 4 includes only 25 simulations. \label{ft:25sims}}. In each simulation, the agents are placed in a random starting area and location, and assigned random goal areas. Their routes are computed given the topological graph of the environment. We, furthermore, make sure that the combined routes of the agents satisfy the earlier introduced assumptions on the routes of agents, see Section \ref{sec:claimPolicy}. In all simulations, agents are identical and homogeneous. The same baseline MPC configuration (e.g., solver configuration and actuation constraints) are applied. The baseline parameters are reported in Table~\ref{tab:param}.

\begin{table}[ht!]
    \centering
    \caption{MPC configuration parameters in simulation experiments}
    \begin{tabular}{lc|cl}
    Parameter                  & Symbol                     & value             &           \\
    \hline
    Longitudonal vel. & $[v_{min}, v_{max}]$           & {[}0.0, 0.5{]}  & {[}$m/s${]} \\
    Longitudonal acc. & $[a_{min}, a_{max}]$           & {[}-0.5, 0.5{]} & {[}$m/s^2${]} \\
    Angular vel.      & $[\omega_{min}, \omega_{max}]$ & {[}-0.5, 0.5{]} & {[}$rad/s${]} \\
    Agent radius               & $r_v$                      & 0.15               & {[}m{]}   \\
    Enlarged agent radius               & $r_{soft}$                      & 0.19               & {[}m{]}   \\
    MPC time step              & $\Delta T$                 & 0.20              & {[}s{]}   \\
    IPOPT max\_cpu\_time & & 0.15 & [s]\\
    MPC update frequency & $f$ & 5 & [Hz] \\
    Prediction horizon     & $N_t$                        & 20                & {[}-{]}  \\
    Semantic Horizon & $N_h$  &   4 & {[}-{]} \\
    \end{tabular}
\label{tab:param}
\end{table}

\subsection{Validation Metrics}
The behaviour of the agents given the proposed method is compared, in a simulation environment, to the behaviour given the methods in \cite{devos2023automatic}, and a no cooperation baseline.\\
The first set of comparison metrics describes the successful completion of the agents' tasks, and thus includes the percentage of successful task completions (succ.), and the percentage of runs which encounter the MPC solver returning infeasible solutions (inf.) and/or in which collisions between agents occur (coll.). The second set of metrics analyzes the scalability of the approach, and include the average number of cooperating agents ($|\mathbb{F}|$), the computation time of the MPC controller ($T_{mpc}$) and the total computation time of one full control cycle ($T_{total}$). Finally we report on the task performance through analyzing the task completion time ($T_1$) of the agents across the different methods.
\subsection{Results}

The successful completion of the tasks of agents is reported on in Table \ref{tab:successration}. The simulation results show that the proposed method is able to solve all the generated navigation problems. The method in \cite{devos2023automatic}, however, fails in a significant number of simulations. It is observed collisions occur in part of the simulations, which can, at least partially, be explained by the limit on CPU time set in the IPOPT configuration in Table \ref{tab:param}. Apart from the simulations in which the solver returns an infeasible status, partly due to the increasing complexity of the optimisation problems, the agents are not able to successfully complete all scenarios, due to getting stuck when meeting another agent in narrow spaces.

\begin{table}[ht!]
    \centering
    \caption{Percentage of successful task completions, infeasibilites and collisions for the method described in this paper, for the method in \cite{devos2023automatic}, and the no cooperation method.}
    \begin{tabular}{c|ccc}
        & \textbf{This Paper} & \cite{devos2023automatic} & \textbf{Never} \\ 
     {[}\%{]}    & [succ., inf., coll.] & [succ., inf., coll.] & [succ., inf., coll.] \\
         \hline
    1 & {[} 100, 0, 0 {]} &  {[} 92, \phantom{1}2, \phantom{1}0 {]} &    {[} 90, 0, 10 {]} \\
    2 & {[} 100, 0, 0 {]} &  {[} 48, \phantom{1}2, 10 {]} &    {[} 48, 0, 52 {]} \\
    3 & {[} 100, 0, 0 {]} &  {[} 18, 12, 33 {]} &   {[} 22, 0, 78 {]} \\
    4 & {[} 100, 0, 0 {]} &  \phantom{1}{[} \phantom{1}4, 16, 48 {]}\footref{ft:25sims} &              {[} \phantom{1}8, 0, 92 {]} \\
\end{tabular}
\label{tab:successration}
\end{table}

With regards to scalability, the MPC and total computation times are reported in Table \ref{tab:tmpc} and \ref{tab:ttotal} respectively. The average amount of agents in each cooperating group is reported on in Table \ref{tab:flocksize}.  The proposed method demonstrates that, in contrast to \cite{devos2023automatic}, MPC computation times scale similarly to those of the no-cooperation baseline, as both solve single-agent MPC problems. Additionally, the overhead on the total computation of the claim policy, across the tested scenarios, remains on average within $11 \ [ms]$ of the no-cooperation baseline. This while \cite{devos2023automatic}  shows that computation times scale badly as the number of agents increases .
The increase in successful task completion performance, and the reduction in computation time, of our algorithms comes, however, at the expense of task completion time. In fact, our proposed method takes on average approximately 12\% more time. In the third scenario this increases to 30\%.

\begin{table}[t]
    \centering
    \caption{MPC computation time, in milliseconds, of completed runs of each scenario for the method described in this paper, for the method in \cite{devos2023automatic}, and the no cooperation method.}
    \begin{tabular}{c|cc|c}
    $T_{mpc}$     & \textbf{This Paper} & \cite{devos2023automatic} & \textbf{Never} \\ 
     {[}ms{]}    & [min, avg, max] & [min, avg, max] & [min, avg, max] \\
         \hline
    1 & {[} 30, 53, 166 {]} &  {[} 29, \phantom{1}74, \phantom{1}203 {]}   & {[} 30, 52, 110 {]} \\
    2 & {[} 31, 61, 173 {]} &  {[} 30, \phantom{1}97, \phantom{1}326 {]}   & {[} 32, 58, 130 {]} \\
    3 & {[} 31, 71, 195 {]} &  {[} 30, 142, \phantom{1}891 {]}             & {[} 33, 66, 147 {]} \\
    4 & {[} 29, 74, 173 {]} &  {[} 27, 186, 1415 {]}             & {[} 30, 69, 161 {]} \\
\end{tabular}
\label{tab:tmpc}
\end{table}

\begin{table}[t]
    \centering
    \caption{Total computation time, in milliseconds, of completed runs of each scenario for the method described in this paper, for the method in \cite{devos2023automatic}, and the no cooperation method.}
    \begin{tabular}{c|cc|c}
    $T_{Total}$     & \textbf{This Paper} & \cite{devos2023automatic} & \textbf{Never} \\ 
     {[}ms{]}    & [min, avg, max] & [min, avg, max] & [min, avg, max] \\
         \hline
    1 & {[} 39, \phantom{1}66, 173 {]}  &  {[} 45, 119, \phantom{1}651 {]}     & {[} 39, \phantom{1}64, 199 {]} \\
    2 & {[} 42, \phantom{1}84, 218 {]}  &  {[} 45, 174, 1461 {]}    & {[} 43, \phantom{1}76, 216 {]} \\
    3 & {[} 43, 104, 224 {]}             &  {[} 49, 315, 3326 {]}    & {[} 45, \phantom{1}89, 211 {]} \\
    4 & {[} 46, 119, 254 {]}            &  {[} 59, 424, 3282 {]}     & {[} 45, 108, 273 {]} \\
\end{tabular}
\label{tab:ttotal}
\end{table}

\begin{table}[ht!]
    \centering
    \caption{Average number of coordinating agents in the largest group for the method described in this paper, for the method in \cite{devos2023automatic}, and the no cooperation method.}
    \begin{tabular}{c|ccc}
      |$\mathbb{F}$|   & \textbf{This Paper} & \cite{devos2023automatic} & \textbf{Never} \\ 
         \hline
    1 &  1.93  &  1.59 &  1.0 \\
    2 &  3.25 &   2.27&1.0  \\
    3 &  4.84 &   3.30  &  1.0 \\
    4 &  5.25 &  4.23  &  1.0  \\
\end{tabular}
\label{tab:flocksize}
\end{table}

\begin{table}[t]
    \centering
    \caption{Task completion time, in seconds, of completed runs of each scenario for the method described in this paper, for the method in \cite{devos2023automatic}, and the no cooperation method.}
    \begin{tabular}{c|cc|c}
    $T_1$     & \textbf{This Paper} & \cite{devos2023automatic} & \textbf{Never} \\ 
     {[}s{]}    & [min, avg, max] & [min, avg, max] & [min, avg, max] \\
         \hline
    1 & {[} \phantom{1}8.0, 12.2, 19.8 {]} &  {[} \phantom{1}6.8, 10.8, 13.8 {]} & {[} \phantom{1}6.8, 10.7, 13.8 {]}\\
    2 & {[} 13.8, 22.1, 33.8 {]} &  {[} 12.8, 18.6, 29.0 {]} & {[} 12.8, 17.4, 22.8 {]}\\
    3 & {[} 16.8, 28.0, 51.4 {]} &  {[} 17.2, 21.3, 32.2 {]} & {[} 12.8, 18.4, 24.8 {]} \\
    4 & {[} 23.0, 41.6, 82.8 {]} &  {[} 26.4, 26.4, 26.4 {]} & {[} 19.8, 26.3, 34.8 {]} \\
\end{tabular}
\label{tab:ttask}
\end{table}

\subsection{Validation on physical system}

To validate the proposed framework on physical hardware, tests are performed with the ROSbot 2 PRO mobile robotic platform. The on-board computer of the robot is running AMCL \cite{fox2001kld} for localization, the presented framework is running on an off-board central desktop computer\footnote{CPU: Intel Core i7-14700K, RAM: 32GB.}. We validate the simulation results with two agents, and present the results both in Figure \ref{fig:experimentalresults} and  in the accompanying video.\footnote{\url{https://youtu.be/ULQ0uxgj9FQ}}. The experiments showed that the general behaviour of the agents generalizes to physical systems, however, it also showed the need to increase the safety radius to $0.3 [m]$ to mitigate the effects of imperfect localisation.

\section{Conclusions}

This paper presented an approach that enables teams of mobile robots to navigate structured dynamic environments while avoiding deadlocks and collisions, both with each other and environmental elements. The proposed algorithms build upon methods from our earlier work, where we proposed the automatic configuration of model predictive controllers using semantic graph world models. By implementing a claim policy, we decentralize decision-making and allow for the avoidance of deadlock situations
Experiments conducted both in simulation and on physical robotic hardware demonstrate the relevance and applicability of the method. A comparative study indicates that the proposed approach results in reduced computation times and higher rates of successful task completion compared to our previous work. However, the conservative nature of the approach leads to an increase in task completion times, up to one-third longer in the tested scenarios, for successfully completed tasks.
Addressing the conservative aspects of the approach is considered future work. Some situations, such as two agents travelling in the same direction, may not (always) require the strict one-agent-per-area constraints considered in this paper. Developing the logic and corresponding MPC formulation to accommodate such situations while maintaining reduced MPC computation times and deadlock avoidance could be an interesting research direction.

\section*{Acknowledgments}
This work has been executed as part of the Semantic navigation for Teams of Open World Robots (TOWR) research project and is part of the ICAI EAISI FAST lab

\printbibliography

\end{document}